# Reducing Artificial Neural Network Complexity:
# A Case Study on Exoplanet Detection

**Sebastiaan Koning, Caspar Greeven and Eric Postma**
Department of Cognitive Science and Artificial Intelligence
Tilburg University, Netherlands

## Abstract

Despite their successes in the field of self-learning AI, Convolutional Neural Networks (CNNs) suffer from having too many trainable parameters, impacting computational performance. Several approaches have been proposed to reduce the number of parameters in the visual domain, the Inception architecture [Szegedy et al., 2016] being a prominent example. This raises the question whether the number of trainable parameters in CNNs can also be reduced for 1D inputs, such as time-series data, without incurring a substantial loss in classification performance. We propose and examine two methods for complexity reduction in AstroNet [Shallue & Vanderburg, 2018], a CNN for automatic classification of time-varying brightness data of stars to detect exoplanets. The first method makes only a tactical reduction of layers in AstroNet while the second method also modifies the original input data by means of a Gaussian pyramid. We conducted our experiments with various degrees of dropout regularization. Our results show only a non-substantial loss in accuracy compared to the original AstroNet, while reducing training time up to 85 percent. These results show potential for similar reductions in other CNN applications while largely retaining accuracy.

## 1   Introduction

Recent advancements in computational technology have made it possible to employ powerful artificial intelligence (AI) techniques to automatically classify large datasets with many features. This branch of AI is known as *representation learning* [LeCun, Bengio & Hinton, 2015], including the fields of machine learning and deep learning. A recent very successful deep learning application is the convolutional neural network (CNN), first theorized by LeCun et al. [1989] and later implemented as LeNet5 [LeCun, Bottou, Bengio & Haffner, 1998]. CNNs have proven to be especially effective in image and video recognition [LeCun et al., 2015]. Many improvements upon LeNet5 have been made since, such as demonstrated by e.g. AlexNet [Krizhevsky, Sutskever & Hinton, 2012], ResNet [He, Zhang, Ren & Sun, 2016] and the Inception Network [Szegedy et al., 2015].

Despite their successes, CNNs pose a considerable computational challenge. Training CNNs requires a lot of computational power and therefore training time due to the large number of trainable parameters that need to be optimized. Because of this, research into CNN application has somewhat shifted from optimizing accuracy towards reducing computational cost while retaining accuracy as much as possible. We will review these efforts in section 2.

This research intends to extend upon the existing effort to make CNNs more efficient with a case study that explores parameter reduction in the domain of signal processing. Specifically, our study focusses on AstroNet [Shallue & Vanderburg, 2018]. AstroNet is a one-dimensional CNN which detects exoplanets from subtle patterns in time-varying stellar brightness signals, so-called *stellar light curves*. The AstroNet architecture uses two input views; a global view representing an aggregation of available stellar light curves on the exoplanet candidate, and a local view that focuses on the transit event, i.e., the dip in brightness caused by the passing of the exoplanet in front of the star. Both views have their own hidden representation and intermediate outputs, which are then concatenated and fed to fully connected layers before the final output is predicted. This is illustrated in Figure 1, showing the global view and its hidden representation on the left, while the local view is shown on the right. Both views are concatenated before being processed by multiple hidden layers to generate a scalar output. It is important to note that, the actual AstroNet architecture contains two and five blocks of convolution and pooling layers for the local and global views, respectively. Each block consists of two convolution layers followed by a pooling layer. In addition, after the concatenation of both views, the architecture has four fully-connected layers that feed into a sigmoid unit.

The best performing AstroNet architecture achieves an accuracy of 96 percent, making it a suitable model for automatically classifying stellar light curves. However, the AstroNet model is quite complex as it consists of almost 30 million trainable parameters, which makes it a computationally intensive model.

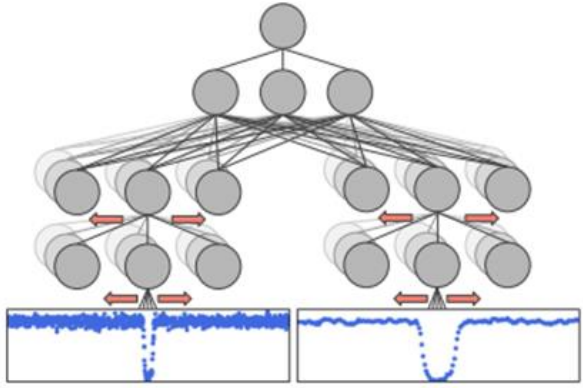

Figure 1: Simplified illustration of the AstroNet architecture with a global and a local view as inputs [Shallue & Vanderburg, 2018]. The arrows indicate the repetition of the localized convolution filters across each input view.

This raises the question whether the complexity of the model could be reduced to limit computational requirements at the cost of only a non-substantial performance loss. The research question central to this case study can therefore be stated as follows: 'Can the complexity of the AstroNet model be reduced, without incurring substantial loss in exoplanet detection performance?'

We propose and examine two methods for complexity reduction in AstroNet to answer this question. The first method comprises a tactical reduction in the number in convolutional and max pooling layers, as well as reducing the number of fully connected layers and their unit size. The input remains unchanged in this method. The second method uses a reduced representation of the global view generated by a Gaussian pyramid [Burt & Adelson, 1983]. Gaussian pyramids have seen several successful applications in image processing to benefit deep learning, e.g., in Sajda, Spence, Hsu & Pearson [1995] and He, Zhang, Ren & Sun [2014]. We experiment with varying degrees of dropout regularization in both experimental methods and the original AstroNet to regularize the induced models. The original model by Shallue & Vanderburg [2018] is used as a baseline for accuracy and training time.

The outline of this paper is as follows. Section 2 provides an overview of related work in parameter reduction in neural networks. This section is followed by section 3 outlining our experimental setup, of which the results can be found in section 4. These results will be discussed in section 5 and concluded upon in section 6, with the latter containing suggestions for future work as well.

## 2 Related Work on Parameter Reduction

The conception of the AlexNet CNN [Krizhevsky et al., 2012], which made use of GPUs to efficiently run a network with 60 million trainable parameters on the ImageNet dataset, made very deep and wide CNNs available for a broad audience. Moreover, it significantly increased performance over other neural networks such as LeNet5. Thereafter, a considerable effort has been made to scale up neural networks even further in the quest to improve model accuracy. Networks containing several billions of trainable parameters were devised, such as a model trained by Coates et al. [2013] containing 11 billion parameters. Large models such as these require immense computational power to train and have large memory requirements, making them unsuitable for storage on smaller machines and rapid transfer to other devices [Han, Pool, Tran & Dally, 2015; Iandola et al., 2016].

As a result of these complications, an increasing amount of research now focuses on reducing the computational load of deep neural networks. One of the most notable ongoing innovations can be found in the work on the Inception network. Its first version was introduced as GoogLeNet, which subjected layer input to computationally cheap 1x1 convolutions in parallel to reduce the number of features before executing convolutions with larger kernel sizes [Szegedy et al., 2015]. This greatly reduced the number of parameters and operations to be performed by the more expensive convolutional layers. Later versions of Inception added batch normalization methods (Inception-v2 and v3) [Ioffe & Szedegy, 2015; Szegedy, Vanhoucke, Ioffe, Shlens & Wojna, 2016], as well as restructuring Inception's architecture in Inception-v4 [Szegedy, Ioffe, Vanhoucke & Alemi, 2017]. Further innovations within Inception were also made by incorporating the concept of residual connections from ResNet [He et al., 2016] within Inception (Inception-ResNet) [Szegedy et al., 2017].

Other notable research by Han et al. [2015] demonstrated pruning of redundant connections in AlexNet during training, thereby reducing the number of trainable parameters within AlexNet by almost a ten-fold to only 6.7 million without accuracy loss. Finally, Iandola et al. [2016] introduced SqueezeNet, which employ *squeeze* layers consisting of 1x1 convolution filters before being passed onto *expand* layers with both 1x1 and 3x3 convolution filters, together called a Fire module. This innovation achieved results comparable to those achieved with AlexNet on the ImageNet dataset while requiring 50 times fewer parameters. The reduction in the number of parameters also greatly benefits model transferability between devices.

Most related work, including those mentioned in this section, focuses on 2D- and 3D-CNNs for image and video recognition. To the best of our knowledge, no directly related work on parameter reduction in 1D-CNNs existed at the time of our review. A few papers propose a 1D-CNN as an efficient alternative for statistical analysis and other machine learning and deep learning tools, e.g., in research by George, Xie & Tam [2018] and by Eren, Ince & Kiranyaz [2018].

## 3 Experimental Setup

This section first outlines the properties of the dataset that was used and how it was preprocessed in subsection 3.1. The configuration of all the networks will be described in subsection 3.2, followed by the description of the experimental procedure and the evaluation methods used in sub-section 3.3. Finally, subsection 3.4 outlines which software and hardware were used.

## 3.1 Dataset and preprocessing

The dataset used comprises all recorded stellar light curves for all Kepler Threshold Crossing Events (TCEs), i.e., all recorded transits resembling an exoplanet transit, in the Kepler Data Release 24 (DR24) [Coughlin et al., 2016]. There are four different training labels for the TCEs in DR24: Planet Candidate (PC), astrophysical false positive (AFP), non-transiting phenomena (NTP) and unknown (UNK). TCEs of the latter category were excluded from our dataset as it is not known whether these transits are of exoplanets or not. All stellar light curves have been obtained from the website of the Mikulski Archive for Space Telescopes [MAST, 2018]. The data format of the stellar light curves is the Flexible Image Transport System (FITS). The dataset contains recorded stellar light curves on 15,740 TCEs.

Except for the extra steps required to generate a reduced representation from a Gaussian pyramid, our preprocessing procedure is equal to the preprocessing procedure used by Shallue & Vanderburg (2018). Below, we provide a brief summary and refer to their paper for a full explanation on the preprocessing. All light curves were read from FITS files, removing any confirmed exoplanet transits in the stellar system. After isolating remaining transits and other non-transit outliers, the light curves were flattened according to fitting a best-fit spline to the remaining data points. The best-fit for the spline was chosen using the Bayesian Information Criterion (BIC). All light curves were then folded into vectors of equal length, generating 'global view' vectors of length 2001 and 'local view' vectors of length 201 centering on the transit event. The reduced Gaussian representation vector (further referred to as 'Gaussian view') of length 251 was generated from a larger view of size 8004. The vector was obtained by the Gaussian pyramid function from the Scikit-image library for Python, using a 5-level pyramid with a downscaling rate of 2 per level. Lastly, brightness was normalized to have a median of 0 and a minimum value of -1. The vectors were saved in two sets of ten equal randomized parts called *shards*. The first set contained the original global and local view inputs, while the second set contained the Gaussian view input. Eight (80% of the data) were used for training, one for validation during training (10%) and one for testing (10%), meaning each shard contained 1,574 TCEs. The test shard contained 360 planet candidates and 1,214 false positives (combination of AFPs and NTP).

## 3.2 Network configurations

This section contains the network configurations for the two experiments. Both configurations share a large number of hyperparameters, outlined below. Differences in configurations are thereafter discussed for each experiment. All models use the Adam optimization algorithm [Kingma & Ba, 2014] with a learning rate $\alpha = 10^{-5}$ and momentum parameters of $\beta_1 = 0.9$, $\beta_2 = 0.999$ and $\epsilon = 10^{-8}$, which are the same parameters that Shallue & Vanderburg [2018] used in their experiments. All layers use the ReLU activation function. All experiments were conducted with varying dropout rates between 0 and 0.3, with steps of 0.1 to determine optimal dropout rate. If dropout was used, it was applied to the fully connected layers only as dropout tends to be less effective for convolutional layers. An overview of the network configurations per experiment can be seen in Table 1, outlining per experiment the number of blocks and the number of convolutional and max pooling layers in each block for the global and, if applicable, local view. Thereafter, the number and sizes of the fully connected layers after convolution (and concatenation if needed) is listed as 'Post-conv' in the view column.

**Baseline**

As stated in the introduction, the baseline configuration comprises five blocks of two convolutional layers and one max pooling layer for the global view, as well as two blocks of two convolutional layers and one max pooling layer for the local view. All convolutional layers have a kernel size of 5. Both views start with 16 feature maps resulting from convolution and are max pooled in each layer using a stride of 2. Pooling window sizes for the global and local views are 5 and 7, respectively. The resulting feature maps are then concatenated and fed to four fully connected layers of size 512, then outputting a prediction using a sigmoid unit.

**Method 1: Decreased Depth Network**

The first method decreases the depth of the original model. The resulting model, the Decreased Depth Network (DDN), reduces the number of blocks of convolutional and max pooling layers for the global view from five to three. Moreover, each block contains only one instead of two convolutional layers before a max pooling layer. The reduction is incorporated in both the global and local views.

| Experiment | View (length) | Blocks | Layers in block |
|---|---|---|---|
| Baseline | Global (2001) | 5 | 2 convolutional layers 1 max pooling layer |
| | Local (201) | 2 | 2 convolutional layers 1 max pooling layer |
| | Post-conv | 1 | 4 FC layers (size 512) |
| DDN | Global (2001) | 3 | 1 convolutional layer 1 max pooling layer |
| | Local (201) | 2 | 1 convolutional layer 1 max pooling layer |
| | Post-conv | 1 | 2 FC layers (size 128) |
| DDMSN | Global (251) | 2 | 2 convolutional layers 1 max pooling layer |
| | Post-conv | 1 | 4 FC layers (size 512) |

Table 1 – Overview of network configurations per experiment

All convolutional layers have a kernel size of 5. Both views start with 16 feature maps resulting from convolution and are max pooled in each layer using a stride of 2. Pooling window size for the global view is 5 and for the local view 7. The resulting feature maps are then concatenated and fed to two, rather than four, fully connected layers of size 128, then outputting a prediction using a sigmoid unit.

**Method 2: Decreasing Depth and Multi-Scale Network**

The second method combines decreasing the number of layers within AstroNet with a reduced representation of the global view generated by a Gaussian pyramid [Burt & Adelson, 1983] called the Gaussian view. Applied to the lightcurve, the Gaussian pyramid decomposes the temporal high-resolution signal into a number of successively lower-resolution copies, each one associated with a level of the pyramid. In our experiments, we employed a 5-level pyramid. Only the final level was used as input for the network. This method, called Decreasing Depth and Multi-Scale Network (DDMSN) also sees a reduction in convolutional and max pooling layers, while the number and size of the fully connected layers remains the same. The Gaussian view is the only input view and is subjected to two collections of two convolutional layers and one max pooling layer, followed by four fully connected layers of size 512 before prediction. All convolutional layers have a kernel size of 5. The initial number of feature maps is 16, while pooling window size is 7 at a stride of 2.

### 3.3 Experimental procedure

The configured models were all trained on the eight training shards and validated during training on the validation shard using the train.py script for 10,000 steps, each step representing one batch of size 64. This translates to approximately 50.8 training epochs by first multiplying batch size and steps, then dividing them by 12,592 training examples. The models were evaluated on prediction accuracy for measuring classification performance, as well as training time and number of trainable parameters as measures for computational intensity.

### 3.4 Used software and hardware

The models were built in TensorFlow, run from an Anaconda Python 3.6.6 environment. The TensorFlow version used is TensorFlow GPU 1.09. To enable GPU acceleration, NVIDIA CUDA 9.0, along with CuDNN 7, was used. The experiments were conducted on a Windows 10 PC with an NVIDIA GeForce GTX 850M GPU, an Intel Core i7-5500U CPU @ 2.40 GHz and 8 GB RAM. Other essential Python packages used in our Anaconda environment include NumPy, AstroPy, Scikit-image and PyDL. We use code that makes use of parts of the original AstroNet code [Shallue & Vanderburg, 2018], supplemented with our own code necessary for the implementation of our research. Links to our code and the original AstroNet code can be found in the Appendix.

## 4 Results

In what follows, first, we present tables of the evaluation metrics described in section 3.3 for each experiment. Then we outline the computational performance of the models by training time, supplemented with the number of trainable parameters per model.

Table 2 provides an overview of the results for each experimental set-up. with optimized dropout and without dropout. The table shows the highest accuracy for the DDN at 95.62 percent, which is 0.6 percent lower compared to the baseline without dropout (i.e. the original AstroNet) at 96.25 percent. For the DDMSN experiment, the highest achieved accuracy is 94.73 percent, which is 1.5 percent lower than the baseline. Interestingly, adding dropout to AstroNet improves its original performance from 0.9625 to 0.9657.

Table 3 specifies the number of trainable parameters per experiment, as well as the training time expressed in minutes (m) and seconds (s) of models without dropout and the average of the three models per experiment that had dropout regularization applied. This is because applying dropout notably increases training time for both experiments. Both the DDN and the DDMSN experiments show a large reduction in training time compared to the baseline, up to 65 and 85 percent respectively compared to the baseline configuration with dropout applied. With dropout applied, the reduction remains 60 and 76 percent respectively. For both experiments, the number of trainable parameters is significantly reduced; from 29.8 million in the original AstroNet to 6.7 million in the DDN experiment and to 5.3 million in the DDMSN experiment.

| Experiment | Accuracy | Dropout |
|---|---|---|
| Baseline | 0.9657 | 0.2 |
| Baseline no dropout | 0.9625 | 0.0 |
| DDN | 0.9562 | 0.3 |
| DDN no dropout | 0.9492 | 0.0 |
| DDMSN | 0.9473 | 0.1/0.2* |
| DDMSN no dropout | 0.9454 | 0.0 |

* Both dropout rates yield equal performance

Table 2 – Accuracy and optimal dropout rate (if applied) per experiment. The AstroNet architecture corresponds to "Baseline no dropout".

| Experiment | Parameters (millions) | Train time no dropout | Train time with dropout |
|---|---|---|---|
| Baseline | 29.8 | 39m50s | 41m20s |
| DDN | 6.7 | 14m37s | 16m39s |
| DDMSN | 5.3 | 5m49s | 10m27s |

Table 3 – Number of trainable parameters and model training time with and without dropout per experiment

## 5 Discussion

We conducted experiments with a Decreased Depth Network, featuring tactical layer and layer size reduction, and experiments with a Decreased Depth Multi-Scale Network, which made use of one reduced input representation generated by a Gaussian pyramid and a reduction in convolutional and max pooling layers. The experiments resulted in around 5 times fewer trainable parameters for the DDN experiment and around 6 times fewer for the DDMSN experiment.

A closer look at the results of the DDN experiment shows that accuracy has only dropped 0.6 percent as compared to the original AstroNet setup (95.62% versus 96.25%), provided a dropout rate of 0.3 is applied. Meanwhile, training time was reduced by 60 percent. Implementing dropout does increase training time slightly over our implementation without dropout, as the implementation without dropout reduces training time by 65 percent. However, the additional reduction in accuracy loss makes applying dropout a worthwhile investment as training time reduction remains significant. The same is also true for the original AstroNet, in which a dropout rate of 0.2 slightly increases accuracy at 0.3 percent.

The results of the DDMSN experiment show an even larger reduction in training time of 85 percent compared to the baseline and is almost three times faster than the setup of the DDN experiment, provided no dropout is added. Though lower than in the DDN experiment, classification performance remains at an acceptable level of 94.54 percent given the reduction in training time. Adding a dropout rate of 0.1 or 0.2 increases accuracy in this setup by 0.2 percent, however at the cost of an almost two-fold increase in training time, making this addition less attractive as compared to the DDN experiment.

After having completed our experiments, we became aware of recent work by Ansdell et al. [2018] who improved AstroNet by incorporating domain knowledge in a model called Exonet. To this end, Exonet features an additional input view containing so-called centroid information of a TCE, that helps to reduce the false-positive rate of exoplanet detection, as well as a number of relevant stellar parameters. Ansdell et al. [2018] also improved the training data for AstroNet and added random Gaussian noise to the training input to prevent overfitting. The Exonet model achieved an accuracy of 97.5 percent, a significant improvement over the original AstroNet. Additionally, the authors proposed Exonet-XS, which featured a similar tactical reduction of the network complexity as proposed by us as well as "[introducing] global max pooling at the output of each convolutional column" (Ansdell et al. [2018], p. 5). Despite having only 0.07% of the original AstroNet parameters, Exonet-XS still performed very well thanks to the improvements made with domain knowledge, achieving an accuracy of 96.6 percent.

Our results supplemented with those of Ansdell et al. [2018] are not only important for the AstroNet model but show the potential of parameter reduction for other tasks involving one-dimensional data. This may be of relevance in cases where computational resources are scarce. In the wider perspective of attempts to reduce the parameter complexity of deep learning models, our work provides another example of the benefits of simplifying neural network architectures.

## 6 Conclusions and Future Work

The research question central to our study was: 'Can the complexity of the AstroNet model be reduced, without incurring substantial loss in exoplanet detection performance?' Our experiments featuring a tactical reduction in layers (Decreased Depth Network experiment) and one combining it with a reduced input representation generated by a Gaussian pyramid (Decreased Depth Multi-Scale Network experiment) reduced the number of trainable parameters of AstroNet from 29.8 million to 6.7 and 5.3 million respectively. The first experiment achieved an accuracy of 95.62 percent with a dropout rate of 0.3, which is only 0.6 percent lower than the performance of the original AstroNet, while training time was reduced by 60 percent. The second experiment manages to reduce training time even further (85 percent less than the baseline) at an accuracy of 94.54 percent. While the accuracy in the DDMSN experiment is one percent lower than in the DDN experiment, the near threefold reduction in training time in the DDMSN experiment versus the DDN experiment makes it an efficient choice at the cost of a slight additional drop in accuracy. The optimal approach to be taken between these results is based on computational considerations and the importance given to good performance and model efficiency.

Future work on parameter reduction of AstroNet could see a combination of the ideas behind our work and the ideas behind Exonet-XS [Ansdell et al., 2018]. For example, one could generate a reduced representation using a Gaussian pyramid of the stellar light curve and/or the centroid view of a TCE, which would reduce computational intensity even further and hopefully preserve most of the accuracy. Ideas underlying the successful simplified architectures for image-related 2D tasks, such as the Inception V3 architecture, may be applied to 1D variants. In addition, regularization methods may be feasible, such as the introduction of random Gaussian noise as used by Ansdell et al. [2018], applying dropout to the fully connected layers like in our experiments, or another regularization method such as stochastic pooling [Zeiler & Fergus, 2013] within the convolutional layers. More broadly speaking, future work could focus on reducing parameters in other 1D-CNNs using one or more of our methods in combination with scientific domain knowledge.

### Acknowledgments

We would like to thank NASA for making the data of the Kepler telescope publicly available, as well as Chris Shallue and Andrew Vanderburg for publicly sharing the AstroNet source code with instructions.

# Appendix: Links to source code

Original AstroNet GitHub: https://github.com/google-research/exoplanet-ml

Anonymous link to zip file containing a GitHub fork of the AstroNet source code with our modifications: https://gofile.io/?c=nTk3vK